\pdfoutput=1

\documentclass[11pt]{article}

\usepackage[final]{acl}

\usepackage{times}
\usepackage{longtable}
\usepackage{latexsym}
\usepackage{booktabs}
\usepackage{adjustbox}
\usepackage{multicol}
\usepackage{multirow}
\usepackage{amsmath}
\usepackage{amsfonts}
\usepackage{amssymb}
\usepackage{pdflscape}
\usepackage{philex}
\usepackage{enumitem}
\usepackage{bbm}

\usepackage{hhline}
\usepackage[T1]{fontenc}

\usepackage[utf8]{inputenc}

\usepackage{microtype}

\usepackage{tcolorbox}

\usepackage{inconsolata}

\usepackage{graphicx}

\usepackage{subcaption}

\usepackage{graphicx} 
\usepackage{caption,subcaption}

\title{LLM-Independent Adaptive RAG: \\ Let the Question Speak for Itself}

\author{
 \textbf{Maria Marina\textsuperscript{2,1}},
 \textbf{Nikolay Ivanov\textsuperscript{1}},
 \textbf{Sergey Pletenev\textsuperscript{2,1}},
 \textbf{Mikhail Salnikov\textsuperscript{2,1}},
 \\
 \textbf{Daria Galimzianova\textsuperscript{4}},
 \textbf{Nikita Krayko\textsuperscript{4}},
 \textbf{Vasily Konovalov\textsuperscript{2,5}},
\\
 \textbf{Alexander Panchenko\textsuperscript{1,2},
 \textbf{Viktor Moskvoretskii\textsuperscript{1,3}}}
\\
 \textsuperscript{1}Skoltech,
 \textsuperscript{2}AIRI,
 \textsuperscript{3}HSE University,
 \textsuperscript{4}MTS AI,
 \textsuperscript{5}MIPT
\\
\href{mailto:Maria.Marina@skol.tech}{\{Maria.Marina}, 
\href{mailto:A.Panchenko@skol.tech}{A.Panchenko}, 
\href{mailto:Mikhail.Salnikov@skol.tech}{Mikhail.Salnikov\}}@skol.tech
}

\begin{document}
\maketitle
\begin{abstract}
Large Language Models~(LLMs) are prone to hallucinations, and Retrieval-Augmented Generation (RAG) helps mitigate this, but at a high computational cost while risking misinformation. Adaptive retrieval aims to retrieve only when necessary, but existing approaches rely on LLM-based uncertainty estimation, which remain inefficient and impractical.
In this study, we introduce lightweight LLM-independent adaptive retrieval methods based on external information. We investigated 27 features, organized into 7 groups, and their hybrid combinations. We evaluated these methods on 6 QA datasets, assessing the QA performance and efficiency. The results show that our approach matches the performance of complex LLM-based methods while achieving significant efficiency gains, demonstrating the potential of external information for adaptive retrieval.  

\end{abstract}

\section{Introduction}

Large Language Models~(LLMs) excel in tasks like question answering~(QA)~\cite{DBLP:conf/emnlp/Yang0ZBCSM18, DBLP:journals/tacl/KwiatkowskiPRCP19}, but remain vulnerable to hallucinations~\cite{DBLP:conf/acl/YinZR024, DBLP:journals/corr/abs-2402-10612-rowen}. Retrieval-Augmented Generation~(RAG)~\cite{DBLP:conf/nips/LewisPPPKGKLYR020} mitigates this by incorporating external information, although it introduces risks such as error accumulation~\cite{DBLP:conf/icml/ShiCMSDCSZ23} and external hallucinations~\cite{DBLP:journals/corr/abs-2402-10612-rowen}. 

Adaptive retrieval techniques~\cite{moskvoretskii2025adaptive, DBLP:journals/corr/abs-2402-10612-rowen, DBLP:conf/naacl/JeongBCHP24} aim to balance LLM knowledge with external resources by estimating uncertainty to decide whether retrieval is needed.

However, existing methods primarily frame this task as uncertainty estimation based on LLM internal states or outputs, leading to significant computational overhead. This can offset the efficiency gains from reduced retrieval calls and limit practicality, especially with larger models.

In this study, we address this issue by introducing LLM-independent adaptive retrieval methods that leverage external information, such as entity popularity and question type. Our methods achieve comparable quality while being significantly more efficient, eliminating the need for LLMs entirely.  

\begin{figure}[t!]
    \centering
    \includegraphics[width=\linewidth]{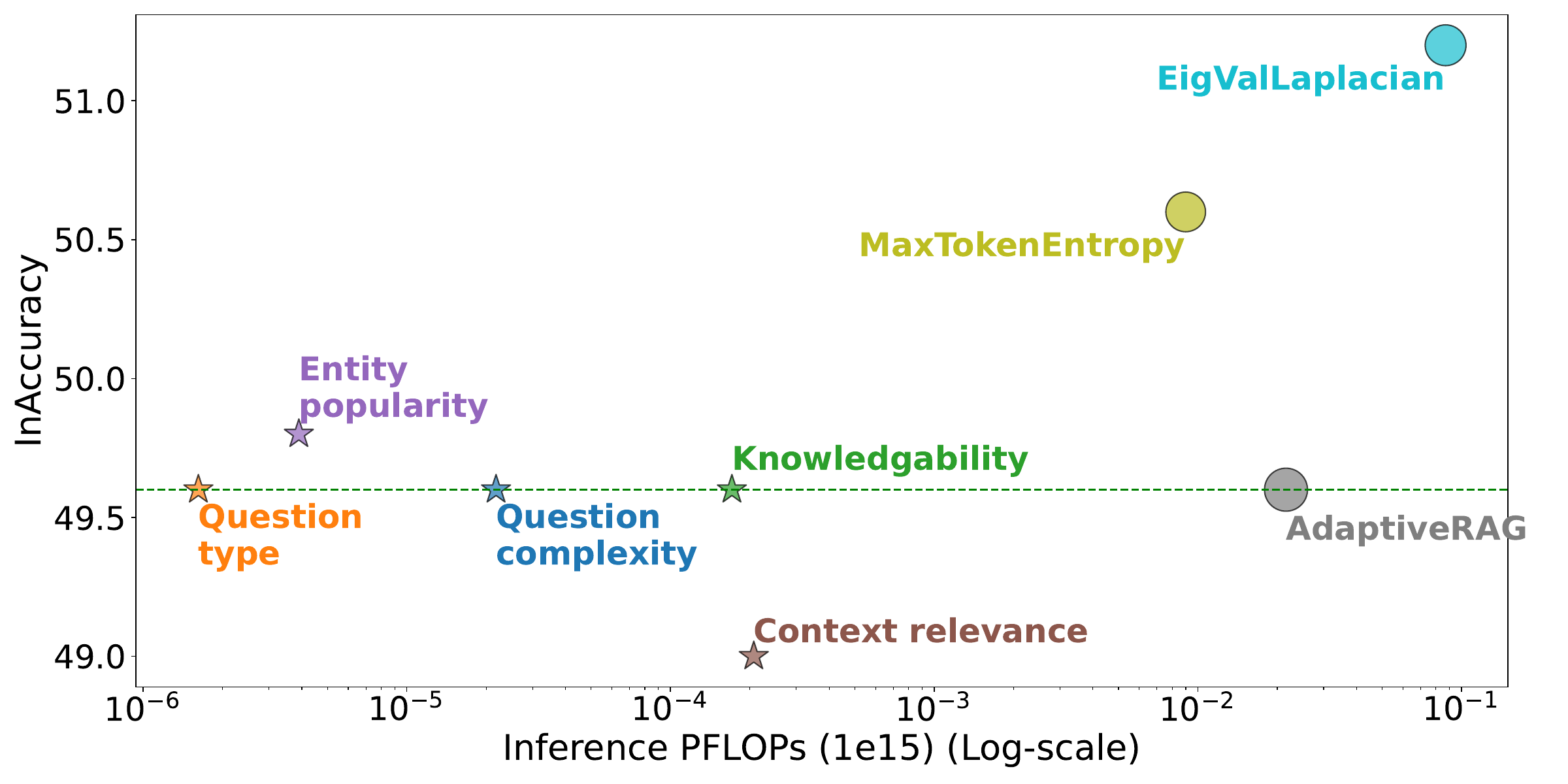}
    \caption{PFLOPs-Inaccuracy trade-off for proposed features vs the most efficient alternative adaptive retrieval methods for the NQ dataset. Radius of the points is proportional to the number of LLM calls. Green dotted line indicate Always RAG approach.
    }
    \label{fig:flops_comparison}
    \vspace{-0.2cm}
\end{figure}

Our evaluation, shown in Figure \ref{fig:flops_comparison}, shows that our proposed features are much more efficient in terms of PFLOPs and LLM calls, with downstream performance comparable to other adaptive retrieval methods.

Our contributions and findings are as follows: 

\begin{enumerate}[itemsep=0.3pt,topsep=0.5pt]

\item We introduce 7 groups of lightweight external information features, encompassing 27 features, for LLM-independent adaptive retrieval.  
\item Our approach significantly improves efficiency by eliminating the need for LLM-based uncertainty estimation while maintaining QA performance.
\item We show that our methods outperform uncertainty-based adaptive retrieval methods for complex questions.

\end{enumerate}

We make data and all models publicly available.\footnote{\url{https://github.com/marialysyuk/External_Adaptive_Retrieval}}

\section{Related Work}


\textbf{Adaptive Retrieval-Augmented Generation} reduces unnecessary retrievals by determining whether external knowledge is needed. This decision can be based on LLM output~\cite{DBLP:conf/acl/TrivediBKS23}, consistency checks~\cite{DBLP:journals/corr/abs-2402-10612-rowen}, internal uncertainty signals~\cite{jiang-etal-2023-active,DBLP:conf/acl/SuTA0024,DBLP:journals/corr/abs-2406-19215}, or trained classifiers~\cite{DBLP:conf/naacl/JeongBCHP24}.

\textbf{External Information} methods can enhance retrieval, such as integrating knowledge graphs and the popularity of the entity. KG structures have been incorporated into LLM decoding to enable reasoning on graphs for more reliable answers~\cite{luo2024graphconstrainedreasoningfaithfulreasoning}. Popularity and graph frequency improve retrieval efficiency, as shown in LightRAG and MiniRAG, which prioritize frequently accessed entities and relationships~\cite{guo2024lightragsimplefastretrievalaugmented, fan2025miniragextremelysimpleretrievalaugmented}. Graph-based features, including entity properties~\cite{DBLP:journals/corr/abs-2409-15902}, popularity~\cite{mallen-etal-2023-trust}, and structural attributes~\cite{salnikov-etal-2023-large}, have also been shown to be effective in QA systems.


\section{Methods}

Our baselines include the following adaptive retrieval methods:

\textbf{Adaptive RAG} uses a T5-large-based classifier to determine whether retrieval is needed~\cite{DBLP:conf/naacl/JeongBCHP24}.
\textbf{FLARE} triggers retrieval when token probability falls below a threshold~\cite{jiang-etal-2023-active}.
\textbf{DRAGIN} estimates uncertainty based on token probabilities and attention weights, excluding stopwords~\cite{DBLP:conf/acl/SuTA0024}.
\textbf{Rowen} relies on consistency checks across languages and models to trigger retrieval~\cite{DBLP:journals/corr/abs-2402-10612-rowen}.
\textbf{SeaKR} monitors internal state consistency to trigger retrieval, re-ranking snippets to reduce uncertainty~\cite{DBLP:journals/corr/abs-2406-19215}.
\textbf{EigValLaplacian} assesses uncertainty using graph features based on pairwise consistency scores~\cite{lin2023generating}.
\textbf{Max Token Entropy} measures uncertainty by aggregating the maximum entropy of token distributions~\cite{fomicheva2020unsupervised}.
\textbf{Hybrid\textsubscript{UE}} includes 5
uncertainty features relevant to the task~\cite{moskvoretskii2025adaptive}: Mean Token Entropy, Max Token Entropy, SAR, EigValLaplacian, Lex-Similarity.

\subsection{External Information Methods}

In this section, we describe the proposed external information methods for adaptive retrieval. Each group may contain multiple features used to train a classifier to predict retrieval needs, following~\citet{moskvoretskii2025adaptive,DBLP:conf/naacl/JeongBCHP24}.

\textbf{Graph} features capture information about the entities in question from a knowledge graph, including the minimum, maximum, and mean number of triples per subject and object, where the subject or object corresponds to an entity from the question.


\textbf{Popularity} features include the minimum, maximum, and mean number of Wikipedia page views per entity in the question.


\textbf{Frequency} features include the minimum, maximum and mean frequencies of entities in a reference text collection\footnote{\url{https://www.inf.uni-hamburg.de/en/inst/ab/lt/resources/data/depcc.html}}, along with the frequency of the least common n-gram in the question.

\begin{table*}[ht!]
    \centering
    \resizebox{\textwidth}{!}{

    \begin{tabular}{lccc|ccc|ccc|ccc|ccc|ccc}
    \toprule
    \multirow{2}{*}{{Method}} & \multicolumn{3}{c|}{NQ} & \multicolumn{3}{c|}{SQuAD} & \multicolumn{3}{c|}{TQA} & \multicolumn{3}{c|}{2Wiki} & \multicolumn{3}{c|}{HotPot} & \multicolumn{3}{c}{Musique} \\
    \cmidrule{2-19}
      & InAcc$\uparrow$ & LMC$\downarrow$ & RC$\downarrow$ & InAcc$\uparrow$  & LMC$\downarrow$ & RC$\downarrow$ & InAcc$\uparrow$ & LMC$\downarrow$ & RC$\downarrow$ & InAcc$\uparrow$ & LMC$\downarrow$ & RC$\downarrow$ & InAcc$\uparrow$ & LMC$\downarrow$ & RC$\downarrow$ & InAcc$\uparrow$ & LMC$\downarrow$ & RC$\downarrow$ \\
    \midrule
    Never RAG & 44.6 & 1.0 & 0.00 & 17.6 & 1.0 & 0.00 & 63.6 & 1.0 & 0.00 & 31.8 & 1.0 & 0.00 & 28.6 & 1.0 & 0.00 & 10.6 & 1.0 & 0.00 \\

    Always RAG & 49.6 & 1.0 & 1.00 & 31.2 & 1.0 & 1.00 & 61.0 & 1.0 & 1.00 & 37.4 & 1.0 & 1.00 & 41.0 & 1.0 & 1.00 & 10.0 & 1.0 & 1.00 \\
    
    \midrule
    \multicolumn{18}{c}{\textit{Multi-Step Adaptive Retrieval}} \\  
    \midrule

    AdaptiveRAG & 49.6 & 2.0 & 0.98 & 28.6 & 2.0 & 0.97 & 62.8 & 1.5 & 0.54 & 45.4 & 5.2 & 2.64 & 41.4 & 4.6 & 2.34 & \textbf{14.0} & 3.6 & 3.63 \\
    DRAGIN & 48.0 & 4.5 & 2.24 & 29.8 & 4.3 & 2.14 & \textbf{66.6} & 4.1 & 2.06 & \textbf{45.6} & 5.8 & 2.92 & \textbf{43.0} & 5.1 & 2.56 & 13.4 & 6.3 & 3.15 \\

    FLARE & 45.0 & 3.1 & 2.07 & 23.8 & 3.1 & 2.08 & 64.8 & 2.1 & 1.39 & 42.4 & 3.9 & 2.85 & 37.2 & 5.1 & 4.07 & 9.0 & 4.1 & 3.10 \\

     Rowen\textsubscript{CM} & 49.4 & 29.5 & 7.27 & 19.6 & 29.2 & 7.20 & 65.6 & 28.7 & 7.12 & 44.4 & 32.9 & 7.87 & 35.6 & 31.9 & 7.70 & 10.4 & 42.1 & 9.52 \\
    
    Seakr & 40.6 & 14.6 & 1.00 & 26.8 & 14.6 & 1.00 & 65.6 & 14.6 & 1.00 & 39.8 & 12.3 & 2.44 & 42.4 & 9.9 & 1.76 & 11.8 & 12.3 & 2.40\\

    \midrule
    \multicolumn{18}{c}{\textit{Uncertainty Estimation}} \\  
    \midrule

    EigValLaplacian & \textbf{51.2} & 1.8 & 0.81 & \textbf{31.4} & 2.0 & 0.10 & 64.0 & 1.3 & 0.26 & 38.4 & 2.0 & 0.98 & 41.0 & 1.9 & 0.91 & 10.2 & 2.0 & 1.00\\

    MaxTokenEntropy & 50.6 & 1.7 & 0.58 & 31.2 & 2.0 & 0.10 & 65.0 & 1.2 & 0.22 & 37.6 & 2.0 & 0.95 & 41.4 & 2.0 & 0.99 & 10.6 & 2.0 & 0.87\\

    Hybrid UE & 50.2 & 1.7 & 0.77 & \textbf{31.4} & 2.0 & 0.98 & 63.8 & 1.3 & 0.27 & 37.4 & 2.0 & 0.98 & 41.2 & 1.9 & 0.94 & 10.6 & 1.8 & 0.75\\

    \midrule
    \multicolumn{18}{c}{\textit{External Features}} \\  
    \midrule
    
    Graph & 49.0 & 1.0 & 0.87 & 30.4 & 1.0 & 0.95 & 63.6 & 1.0 & 0.32 & 35.8 & 1.0 & 0.67 & 40.8 & 1.0 & 0.97 & 10.0 & 1.0 & 1.00\\

    Popularity & 
\textbf{49.8} & 1.0 & 0.92 & \textbf{31.2} & 1.0 & 1.00 & 63.2 & 1.0 & 0.15 & 35.6 & 1.0 & 0.84 & \textbf{41.0} & 1.0 & 0.94 & 10.0 & 1.0 & 0.96\\

     Frequency & \textbf{49.8} & 1.0 & 0.96 & 31.0 & 1.0 & 0.99 & 63.2 & 1.0 & 0.04 & 37.4 & 1.0 & 0.84 & \textbf{41.0} & 1.0 & 0.94 & 10.0 & 1.0 & 0.96\\

    Knowledgability & 49.6 & 1.0 & 0.95 & \textbf{31.2} & 1.0 & 1.00 & 63.0 & 1.0 & 0.28 & \textbf{38.4} & 1.0 & 0.89 & \textbf{41.0} & 1.0 & 1.00 & 9.8 & 1.0 & 0.61\\

    Question type & 49.6 & 1.0 & 0.88 & 30.4 & 1.0 & 0.97 & \textbf{64.0} & 1.0 & 0.29 & 35.6 & 1.0 & 0.74 & 39.6 & 1.0 & 0.88 & 10.0 & 1.0 & 1.00\\

    Question complexity & 49.6 & 1.0 & 1.00 & \textbf{31.2} & 1.0 & 1.00 & 63.6 & 1.0 & 0.10 & 36.8 & 1.0 & 0.94 & \textbf{41.0} & 1.0 & 1.00 & 10.6 & 1.0 & 0.95\\

    Context relevance & 49.0 & 1.0 & 1.00 & 31.0 & 1.0 & 1.00 & 62.8 & 1.0 & 1.00 & 36.0 & 1.0 & 1.00 & \textbf{41.0} & 1.0 & 1.00 & 10.6 & 1.0 & 1.00\\

    \midrule
\multicolumn{18}{c}{\textit{Hybrids with External Features}} \\  
    \midrule

    Hybrid\textsubscript{UFP} & 47.8 & 1.0 & 1.00 & 30.8 & 1.0 & 1.00 & 63.4 & 1.0 & 1.00 & 36.4 & 1.0 & 1.00 & 39.8 & 1.0 & 1.00 & 10.6 & 1.0 & 1.00\\
        
    Hybrid\textsubscript{External} & 46.0 & 1.8 & 1.00 & 30.2 & 2.0 & 1.00 & 63.2 & 1.3 & 1.00 & 37.0 & 2.0 & 1.00 & 39.2 & 1.9 & 1.00 & \textbf{12.2} & 1.8 & 1.00\\

         \midrule
\multicolumn{18}{c}{\textit{Hybrids with Uncertainty and External Features}} \\  
    \midrule
     
     Hybrid\textsubscript{FP}& 48.4 & 1.8 & 1.00 & 31.2 & 2.0 & 1.00 & 64.6 & 1.3 & 1.00 & 37.8 & 2.0 & 1.00 & 41.0 & 1.9 & 1.00 & 12.2 & 1.8 & 1.00\\

      All & 47.6 & 1.8 & 1.00 & 30.8 & 2.0 & 1.00 & 64.2 & 1.3 & 1.00 & 37.8 & 2.0 & 1.00 & 37.8 & 1.9 & 1.00 & 11.2 & 1.8 & 1.00\\

    \midrule
    Ideal & 60.8 & 1.6 & 0.55 & 36.0 & 1.8 & 0.82 & 73.6 & 1.4 & 0.36 & 50.0 & 1.7 & 0.68 & 46.0 & 1.7 & 0.71 & 16.4 & 1.9 & 0.89 \\

    \bottomrule
    \end{tabular}
    }
    \caption{QA Performance of adaptive retrieval and uncertainty methods. `Ideal' represents the performance of a system with an oracle providing ideal predictions for the need to retrieve. `InAcc' denotes In-Accuracy, measuring the QA system's performance. `LMC' indicates the mean number of LM calls per question, and `RC' represents the mean number of retrieval calls per question. The SOTA results are highlighted in bold, as well as the best results for the external methods.}
        \label{tab:main_results_new}

\end{table*}

\textbf{Knowledgability} features assign a score to each entity, reflecting the LLM's verbalized uncertainty about its knowledge. By pre-computing these scores for entities in the Wikidata Knowledge Graph, retrieval decisions can be made without querying the LLM at inference time.


\textbf{Question Type} features include probabilities for nine categories: ordinal, count, generic, superlative, difference, intersection, multihop, comparative, and yes/no.


\textbf{Question Complexity} reflects the difficulty of a question, considering the reasoning steps required.


\textbf{Context Relevance} features include the minimum, maximum and mean probabilities that a context is relevant to the question, along with the context length.

\textbf{Hybrid\textsubscript{External}} includes all external features.

\textbf{Hybrid\textsubscript{$\neg$UFP}}
includes all external features except frequency and popularity, as they are highly correlated with graph features.


\textbf{Hybrid\textsubscript{$\neg$FP}} includes uncertainty and all external features except frequency and popularity.

The details of all methods are described in the Appendix~\ref{sec:external_details}.

\section{Experimental Setup}

In this section, we briefly discuss the implementation details and the evaluation setup.

\subsection{Implementation Details}

We use LLaMA 3.1-8B-Instruct~\cite{dubey2024llama} and the BM25 retriever~\cite{DBLP:conf/trec/RobertsonWJHG94} as the main components of our approach, following~\citet{DBLP:journals/corr/abs-2406-19215, DBLP:conf/naacl/JeongBCHP24, moskvoretskii2025adaptive}.

\subsection{Datasets}

We evaluate on single-hop SQuAD v1.1~\cite{DBLP:conf/emnlp/RajpurkarZLL16}, Natural Questions~\cite{DBLP:journals/tacl/KwiatkowskiPRCP19}, TriviaQA~\cite{DBLP:conf/acl/JoshiCWZ17} and multi-hop MuSiQue~\cite{DBLP:journals/tacl/TrivediBKS22}, HotpotQA~\cite{DBLP:conf/emnlp/Yang0ZBCSM18}, 2WikiMultiHopQA~(2wiki)~\cite{DBLP:conf/coling/HoNSA20} QA datasets to ensure real-world query complexity, following~\citet{DBLP:conf/acl/TrivediBKS23,DBLP:conf/naacl/JeongBCHP24,DBLP:conf/acl/SuTA0024,DBLP:journals/corr/abs-2406-19215}. We use 500-question subsets from the original test sets, as in~\citet{moskvoretskii2025adaptive,DBLP:conf/naacl/JeongBCHP24}.

\begin{figure*}[ht!]
    \centering 
\begin{subfigure}{0.5\textwidth}
  \includegraphics[width=\linewidth]{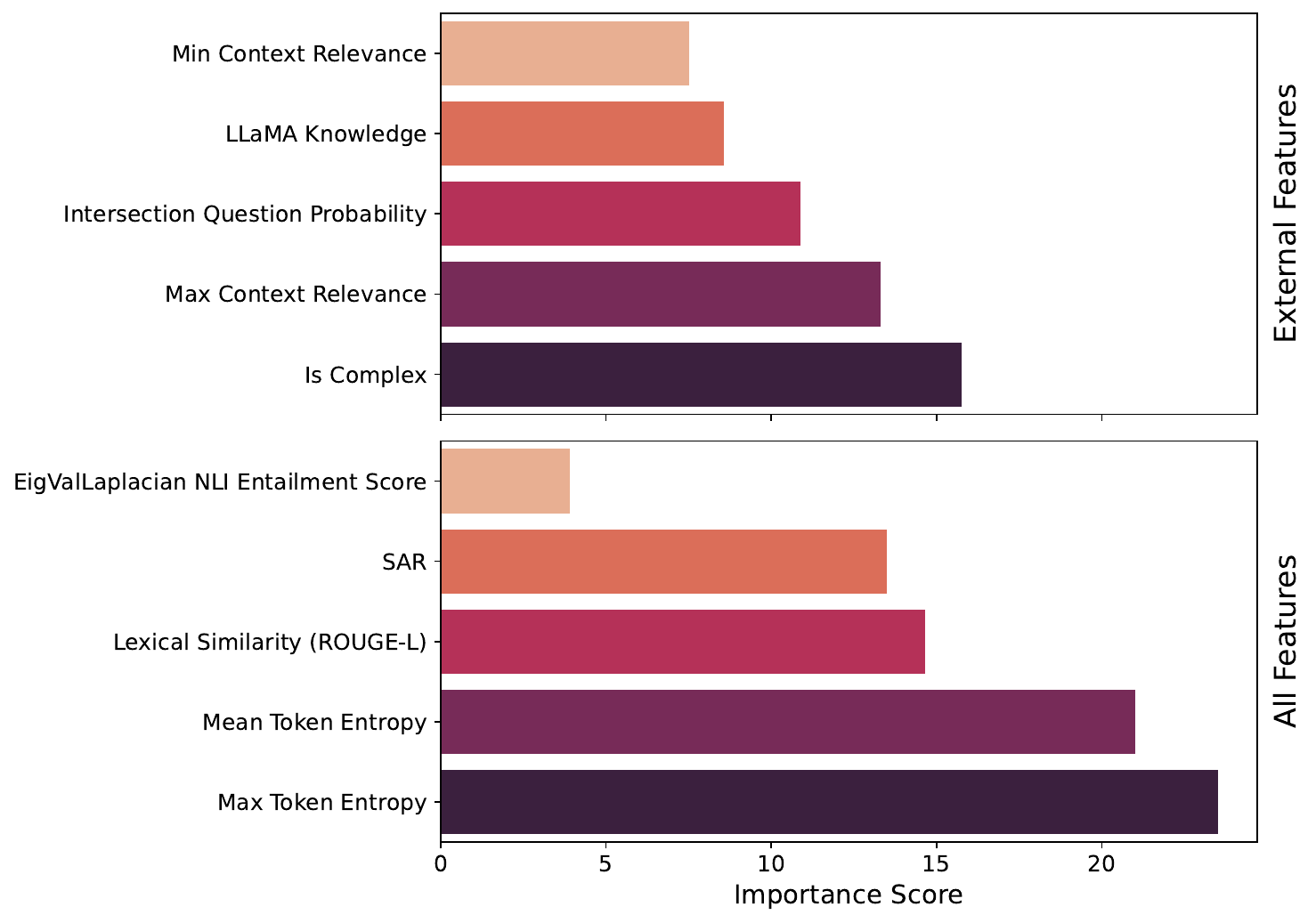}
  \caption{trivia}
  \label{FI_trivia}
\end{subfigure}\hfil 
\begin{subfigure}{0.5\textwidth}
  \includegraphics[width=\linewidth]{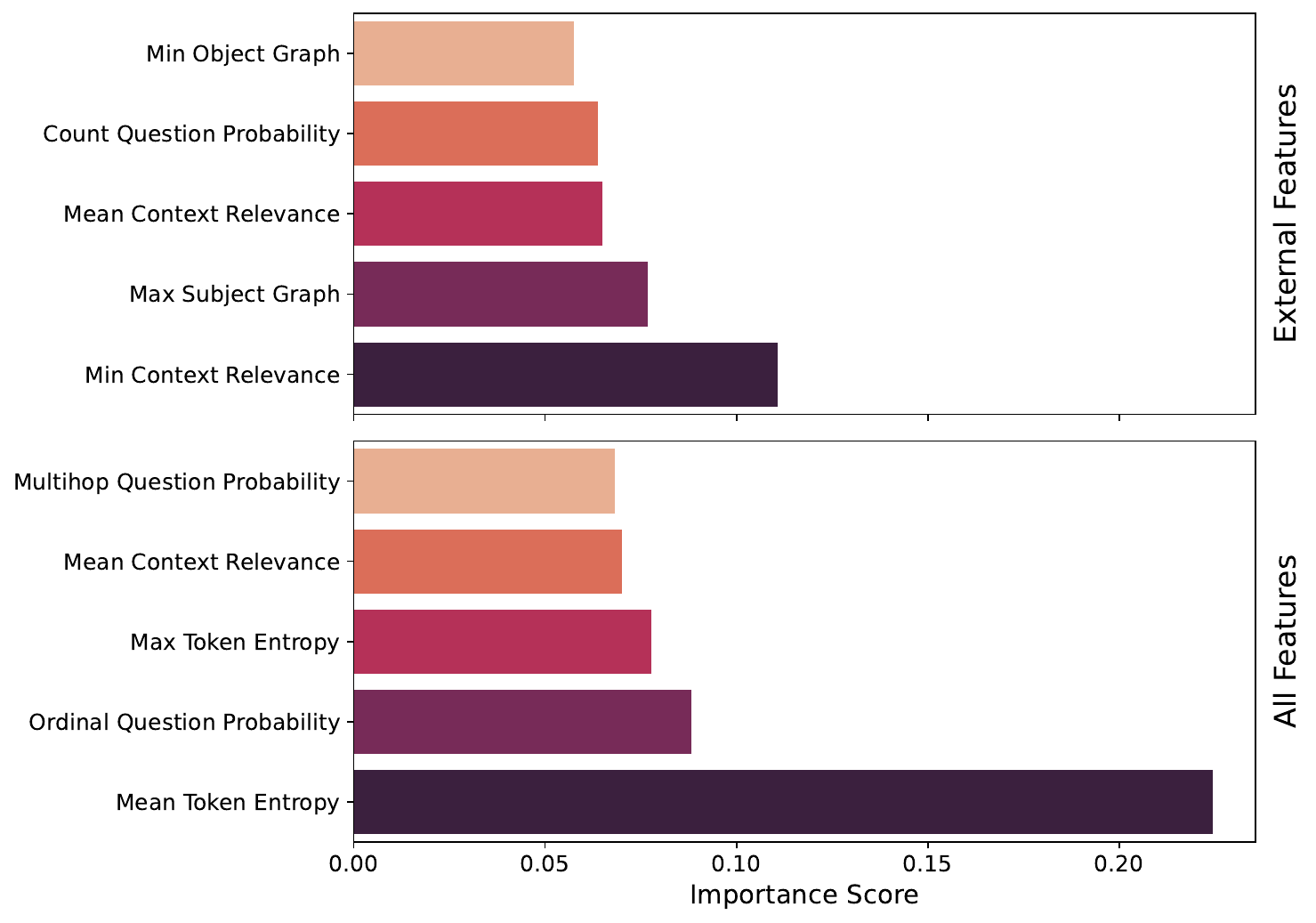}
  \caption{musique}
  \label{FI_musique}
\end{subfigure}\hfil 

\caption{Feature importances for one of the best algorithms for only external features vs all features for TriviaQA (simple) and Musique (complex) datasets. }
\label{fig:FI}
\end{figure*}

\subsection{Evaluation}

We evaluate both the quality and efficiency of the adaptive retrieval system. For quality, we use \textbf{In-Accuracy~(InAcc)}, which measures whether the LLM output contains the ground-truth answer, as it is a reliable metric based on~\citet{moskvoretskii2025adaptive,DBLP:conf/acl/MallenAZDKH23,DBLP:conf/naacl/JeongBCHP24,DBLP:conf/iclr/AsaiWWSH24,DBLP:conf/emnlp/BaekJKPH23}. 

Following~\citet{DBLP:conf/naacl/JeongBCHP24,moskvoretskii2025adaptive}, for efficiency we adopt \textbf{Retrieval Calls (RC)} -- the average number of retrievals per question, and \textbf{LM Calls (LMC)} -- the average number of LLM calls per question, including uncertainty estimation. 
Further details are provided in Appendix~\ref{sec:appendix_technical}.

\section{Results}

In the following sections, we present the results of the end-to-end and UE methods, as well as groups of external features, focusing on downstream performance and efficiency. For comparison, we also include the `Never RAG', `Always RAG', and `Ideal' benchmarks. The `Ideal' benchmark represents the performance of a system with an oracle providing perfect retrieval predictions.


\begin{figure}[t!]
    \centering
    \includegraphics[width=\linewidth]{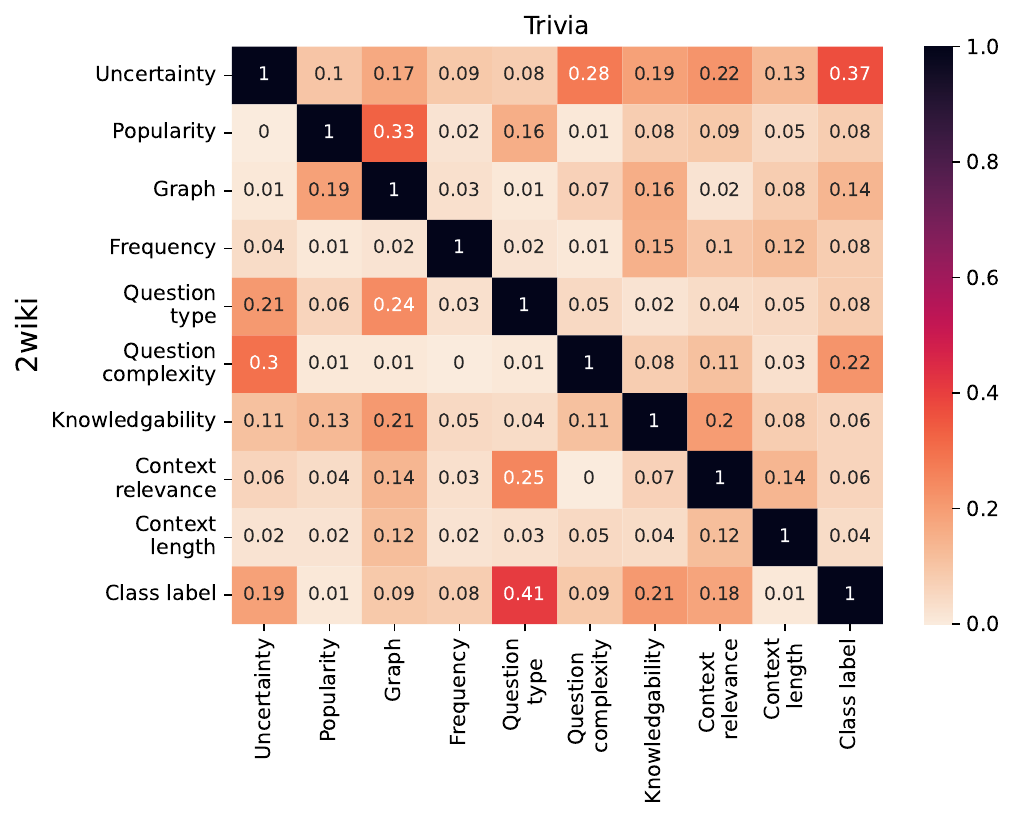}
    \caption{Heatmap of different groups of features for TriviaQA and 2WikiMultiHopQA~(2wiki) datasets. Upper right triangle states for the absolute correlations on the TriviaQA, while down left states for the absolute correlations on the 2WikiMultiHopQA
    }
    \label{fig:corr_all}
\end{figure}

\paragraph{Downstream Performance}

First, we assess whether external methods can replace uncertainty-based approaches. As shown in Table~\ref{tab:main_results_new}, at least one external feature matches the performance of the uncertainty estimation methods for each dataset. Combining external features even increases InAccuracy for the Musique dataset. Compared to Multi-Step Adaptive Retrieval, using only external features yields similar results across all datasets, except for 2wiki.

Second, we examine whether external methods complement uncertainty-based approaches. Our findings show that hybrids with uncertainty features do not outperform any external feature combinations, suggesting that these features are more substitutive than complementary.

\paragraph{Efficiency Performance}

External features significantly reduce LLM calls, addressing a key efficiency bottleneck that worsens with LLM scaling. However, they lead to slightly more conservative behavior with increased Retrieval Calls, though still fewer than Multi-Step approaches.  
Since external information features are pre-computed, no additional LLM calls are required during inference.

\section{Features Reciprocity}

We identify four key aspects that influence adaptive retrieval performance: LLM knowledge (uncertainty features, knowledgability), question type (simple vs. complex reasoning), context relevance (irrelevant context reduces performance), and entity rarity (approximated by entity popularity groups). Figure~\ref{fig:FI} shows that for the simple TriviaQA dataset, the Top-5 features are uncertainty-based, while for complex datasets, question type and context relevance become more important. Thus, relying solely on uncertainty-based features is insufficient for efficient adaptive retrieval.

External features tend to be more substitutive than complementary, as they often exhibit strong correlations despite their differences. As shown in Figure~\ref{fig:corr_all}, for simple questions, uncertainty strongly correlates with graph features, question complexity, knowledgability, and context relevance. For complex questions, uncertainty correlates with question complexity, type, and knowledgability. Heatmaps and feature importances for other datasets could be found in Appendix~\ref{sec:all_pics}.

\section{Conclusion}

In this work, we introduced 7 groups of lightweight external features for LLM-independent adaptive retrieval, improving efficiency by eliminating the need for LLM-based uncertainty estimation while preserving QA performance. Our approach outperforms uncertainty-based methods for complex questions and offers a detailed analysis of the complementarity between uncertainty and external features.

\section*{Limitations}

\begin{itemize}

\item We evaluate model performance using six widely adopted QA datasets. Incorporating a broader range of datasets, particularly those tailored to specific domains, could offer more comprehensive insights and showcase the versatility of our approach.  

 \item Our study focuses on the LLaMA3.1-8B-Instruct model, a top-performing open-source model within its parameter range. Expanding the analysis to additional architectures could further strengthen the generalizability of our results.

\end{itemize}

\section*{Ethical Considerations}

Text retrieval systems can introduce biases into retrieved documents, which may inadvertently steer the outputs of even ethically aligned LLMs in unintended directions. Consequently, developers integrating RAG and Adaptive RAG pipelines into user-facing applications should account for this potential risk.

\bibliography{acl_latex}

\appendix



\clearpage
\onecolumn

\section{External Methods} \label{sec:external_details}

\paragraph{Graph}

Using the BELA entity linking module~\cite{DBLP:journals/corr/abs-2306-08896}, the entities from the question are linked to the corresponding IDs in the Wikidata knowledge graph. Then, for each entity the number of triples where this entity is either an object or a subject is retrieved. Finally, six features are calculated: the minimum/maximum/mean number of triples per subject and object. 

\paragraph{Popularity}

Using the BELA NER module~\cite{DBLP:journals/corr/abs-2306-08896} the entities are retrieved from the question. Then, for each entity the mean amount of views per Wikipedia page is calculated using Wikimedia API\footnote{\url{https://foundation.wikimedia.org/wiki/Api/}} for last year. Finally, there are three features: the minimum/maximum/mean number of views per entity per question.

\paragraph{Knowledgability}

The prompt to the LLaMA 3.1-8B-Instruct model to approximate its interal knowledge:

\begin{figure}[ht!]
   \begin{tcolorbox}[colback=gray!3, colframe=gray!50,fontupper=\itshape]

Answer the following question based 
on your internal knowledge with one or few words. 

If you are sure the answer is accurate and correct, please say `100'. If you are not confident with the answer, please range your knowledgability from 0 to 100, say just number. For example, `40'. 

Question: \{question\}. Answer:

\end{tcolorbox}
\end{figure}

\paragraph{Question type}

Using the train part of the Mintaka dataset~\cite{DBLP:conf/coling/SenAS22}, we train a classifier based on the bert-base-uncased model\footnote{\url{https://hf.co/google-bert/bert-base-uncased}} to predict whether a question belongs to one of the 12 question types: `ordinal', `count', `generic', `superlative', `difference', `intersection', `multihop', `yesno', `intersection', `comparative', `multihop', `yes/no'. As a result, we get twelve probabilities that the question belongs to a certain class. The accuracy classification score on the validation part of the Mintaka dataset is $0.93$.

\paragraph{Question Complexity} is based on N-hop feature from FreshQA~\cite{vu2023freshllmsrefreshinglargelanguage} dataset:
\begin{itemize}
    \item One-hop, where the question is explicit about all the relevant information needed to complete the task, so no additional inference is needed.
    \item Multi-hop, where the question requires one or more additional inference steps to gather all the relevant information needed to complete the task.
\end{itemize}
The dataset consists of 500 training and 100 test examples. As a training model, we used a Distil-bert~\footnote{\url{https://hf.co/distilbert/distilbert-base-uncased}} model. The final F1 score on the test set is $0.82$.

\paragraph{Context relevance}

Each question with one context at a time is passed to the cross-encoder model based on the uncased model of the bert base. A question and a context are passed via the [SEP] token with the additional classification head over the base model. The final probabilities of each context being relevant are aggregated via minimum/maximum/mean across all contexts. Additionally, there is the fourth feature that calculates the context length.

\section{Technical Details} 
\label{sec:appendix_technical}

\paragraph{Train setting.} We conduct all experiments using the LLaMA 3.1-8B-Instruct model with its default generation parameters. The responses generated, with and without the retriever, are sourced from previous studies~\cite{moskvoretskii2025adaptive}, following the AdaptiveRAG framework~\cite{DBLP:conf/naacl/JeongBCHP24}. The baseline results are also adopted from prior work, as we employ the exact same settings and generation configurations.

We implemented classifiers using Scikit-learn~\cite{sklearn_api}, CatBoost~\cite{hancock2020catboost}, and performed hyperparameter tuning using a validation set of $100$ samples randomly selected from the training set, testing with three different random seeds for each dataset. We evaluated seven classifiers: Logistic Regression, KNN, MLP, Decision Tree, CatBoosting, Gradient Boosting, and Random Forest. Data preprocessing involved standard scaling. For the final model, we used a VotingClassifier, combining the two best-performing classifiers from the validation set, each trained with their optimal hyperparameters. Performance was evaluated based on the In-accuracy metric, and the top classifiers were retrained on the full training set with these selected hyperparameters.

\paragraph{Hyperparameters grid.}

\textit{Logistic Regression} :  {C: [0.01, 0.1, 1], solver: [lbfgs, liblinear],
class\_weight: [balanced,  {0: 1, 1: 1}, None], max\_iter: [10000, 15000, 20000]}

\textit{KNN} : {n\_neighbors:  [5, 7, 9, 11, 13, 15], metric: [euclidean, manhattan], algorithm: [auto, ball\_tree, kd\_tree], weights: [uniform, distance]}

\textit{MLP} : {hidden\_layer\_sizes: [(50,), (100,), (50, 50), (100, 50), (100, 100)], activation: [relu, tanh],
solver: [adam, sgd], alpha: [0.00001, 0.0001, 0.001, 0.01],
learning\_rate: [constant, adaptive],
early\_stopping: True,
max\_iter: [200, 500]}

\textit{Decision Tree} : {max\_depth: [3, 5, 7, 10, None], max\_features: [0.2, 0.4, sqrt, log2, None],
criterion: [gini, entropy],
splitter: [best, random]}

\textit{CatBoosting}:  {iterations: [10, 50, 100, 200],
learning\_rate: [0.001, 0.01, 0.05],
depth: [3, 4, 5, 7, 9],
bootstrap\_type: [Bayesian, Bernoulli, MVS]}

\textit{Gradient Boosting}: {
n\_estimators: [25, 35, 50],
learning\_rate: [0.001, 0.01, 0.05],
max\_depth: [3, 4, 5, 7, 9],
max\_features: [0.2, 0.4, sqrt, log2, None]}

\textit{Random Forest}: {
n\_estimators: [25, 35, 50],
max\_depth: [3, 5, 7, 9, 11],
max\_features: [0.2, 0.4, sqrt, log2, None],
bootstrap: [True, False],
criterion: [gini, entropy],
class\_weight: [balanced, {0: 1, 1: 1}, None]}

\section{FLOPs calculation}

\begin{table*}[ht!]
    \centering
    \small
    \begin{tabular}{l|cc}
    \toprule
    \multirow{2}{*}{{Method}} & \multicolumn{2}{c}{NQ} \\
    \cmidrule(lr){2-3}
     & Mean & Upper bound \\
    \midrule
    AdaptiveRAG & 0.0216 & 0.4389\\
    SeaKR & 0.3504 & 2.4548 \\
    DRAGIN & 0.2608 & 1.0129 \\
    FLARE & 0.09699 & 0.9290 \\
    Rowen & 1.865 & 15.9677 \\
    \midrule
    EigValLaplacian & 0.10517 & 0.3291 \\
    MaxTokenEntropy & 0.027116 & 0.22121 \\
    Entity\_popularity & 0.0181238962 & 0.210304 \\
    Is\_complex & 0.0181418 & 0.2082277 \\
    Llama\_know & 0.018291 & 0.22747 \\
    Context\_relevance & 0.018327 & 0.2084429 \\
    Question\_type & 0.01812162 & 0.2073669 \\
    \bottomrule
    \end{tabular}
    \caption{A comparison of FLOPs usage across different methods on the Natural Questions (NQ) dataset. The “Mean” column shows the average PFLOPs ($10^{15}$ FLOPs) per question, while the “Upper bound” column represents the theoretical maximum FLOPs assuming the LLaMA 3.1 8B model (in FP16 precision) runs at 100\% GPU utilization for the entire processing of a single sample. The row labeled “Entity\_popularity” reflects the computational overhead required for graph/popularity/frequency features. It is important to note that for features such as "Entity\_popularity", "Is\_complex", "Llama\_know", "Context\_relevance", "Question\_type" the generation of final answer for a question (after precomputing these features) accounts for more than 99\% of the total FLOPs.}
    \label{tab:main_results}
\end{table*}

To calculate floating-point operations (FLOPs), we used the fvcore~\cite{fvcore_api} library developed by Facebook Research. This library provides a flexible and efficient interface for analyzing the computational complexity of PyTorch models. Specifically, we wrapped our model generation process with the \texttt{FlopCountAnalysis} class, which automatically traces the model forward pass and counts the number of FLOPs for each layer. The theoretical analysis includes an approximate formula to calculate an upper bound per sample:
\[
\text{Total FLOPs} \approx 
    \bigl(\text{Total TFLOPs}\bigr) \times 10^{12} \times (\text{Elapsed Seconds}),
\]
\[
\text{where } \text{Total TFLOPs} = 
    (\text{TFLOPs per GPU}) \times (\text{Number of GPUs}), \quad \text{assuming 100\% utilization.}
\]

\section{Heatmaps and feature importances for all datasets}
\label{sec:all_pics}

\begin{figure*}[ht!]
    \centering 
\begin{subfigure}{0.5\textwidth}
  \includegraphics[width=\linewidth]{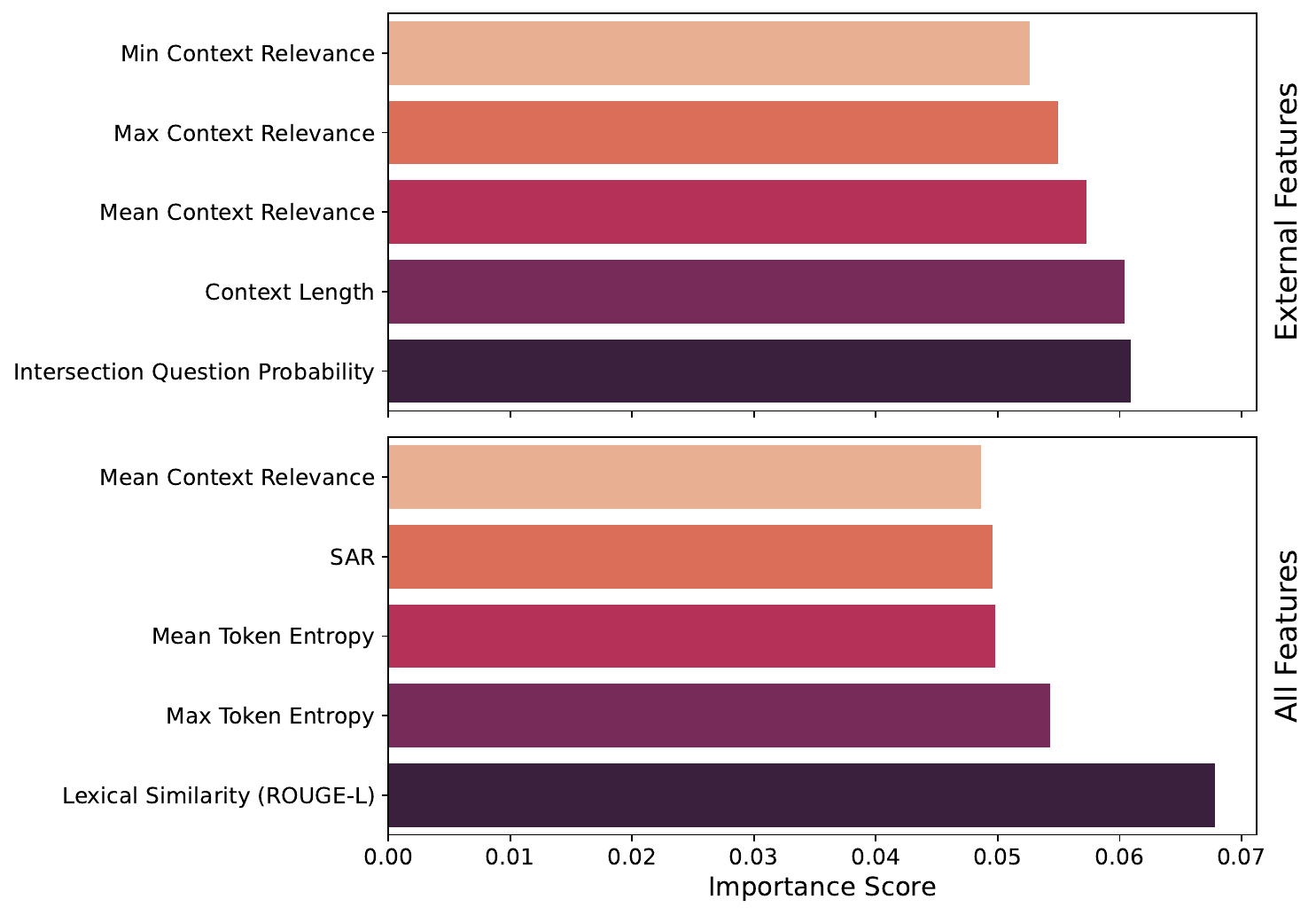}
  \caption{nq}
  \label{FI_nq}
\end{subfigure}\hfil 
\begin{subfigure}{0.5\textwidth}
  \includegraphics[width=\linewidth]{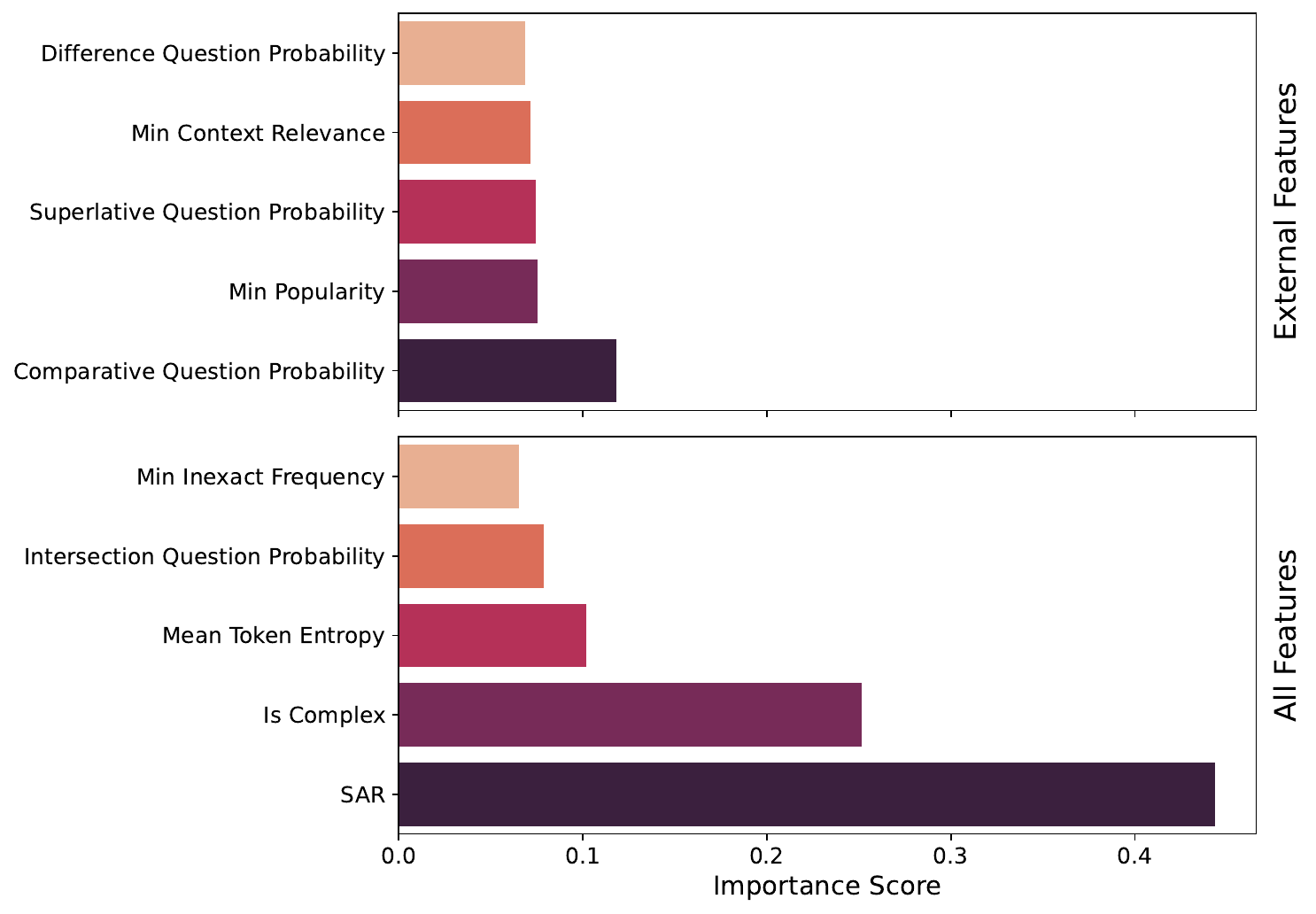}
  \caption{squad}
  \label{FI_squad}
\end{subfigure}\hfil 

\medskip

\begin{subfigure}{0.5\textwidth}
  \includegraphics[width=\linewidth]{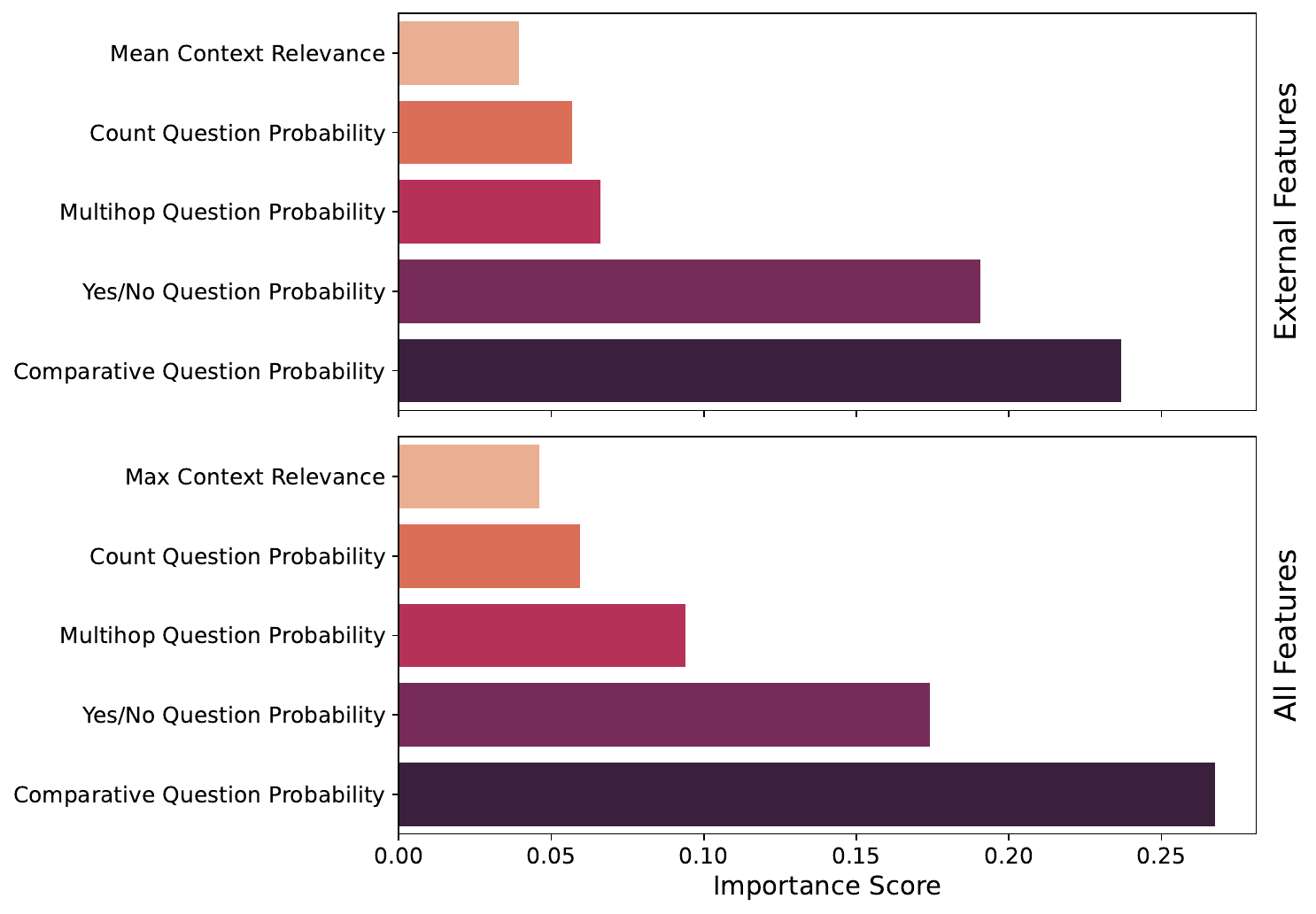}
  \caption{2wikimultihop}
  \label{FI_2wiki}
\end{subfigure}\hfil 
\begin{subfigure}{0.5\textwidth}
  \includegraphics[width=\linewidth]{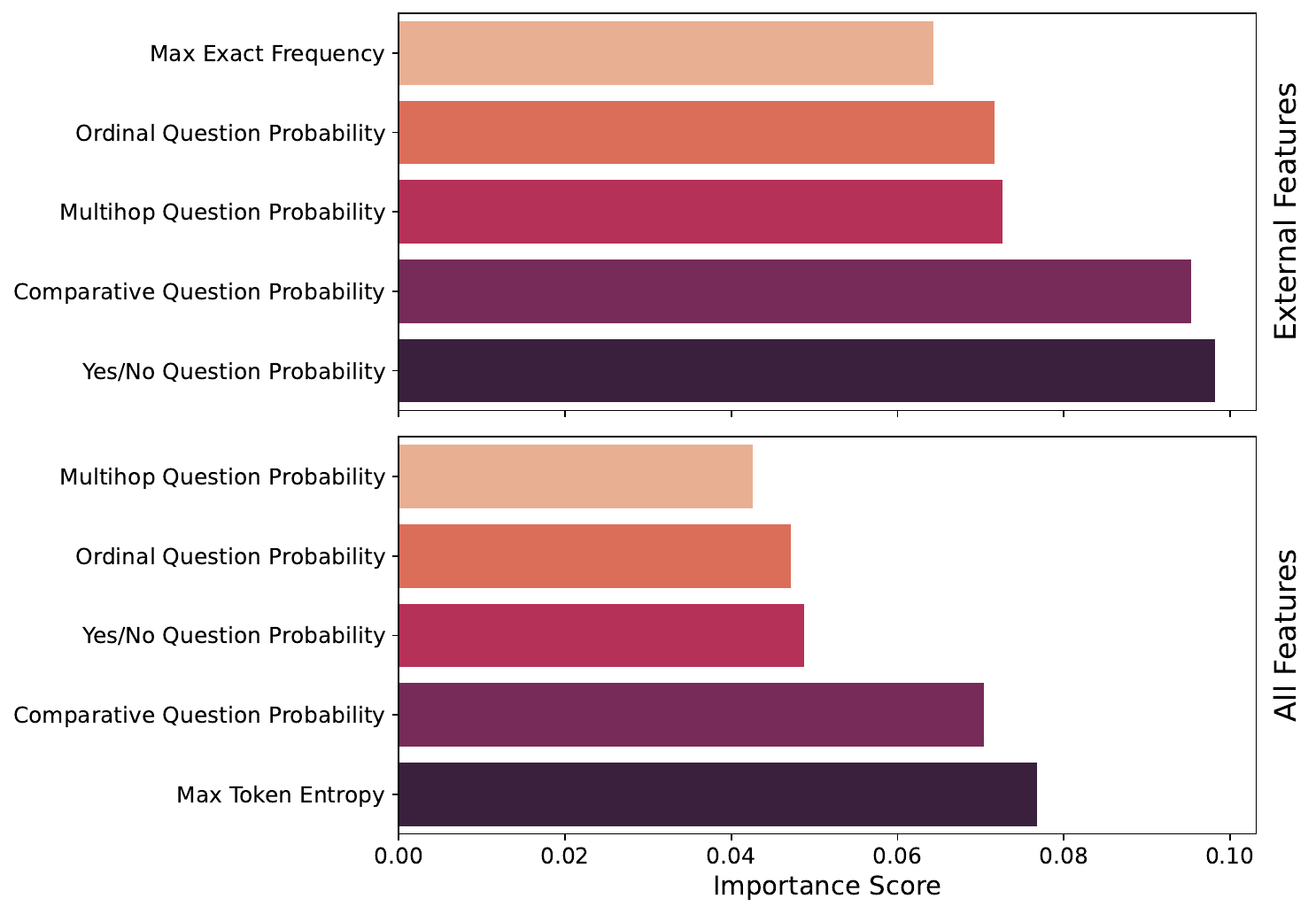}
  \caption{hotpot}
  \label{FI_hotpot}
\end{subfigure}\hfil 

\caption{Feature importances for one of the best algorithms for only external features vs all features for NQ, TriviaQA (simple) and HotpotQA, Musique (complex) datasets. }
\label{fig:FI_full}
\end{figure*}

\begin{figure*}[ht!]
    \centering 
\begin{subfigure}{0.5\textwidth}
  \includegraphics[width=\linewidth]{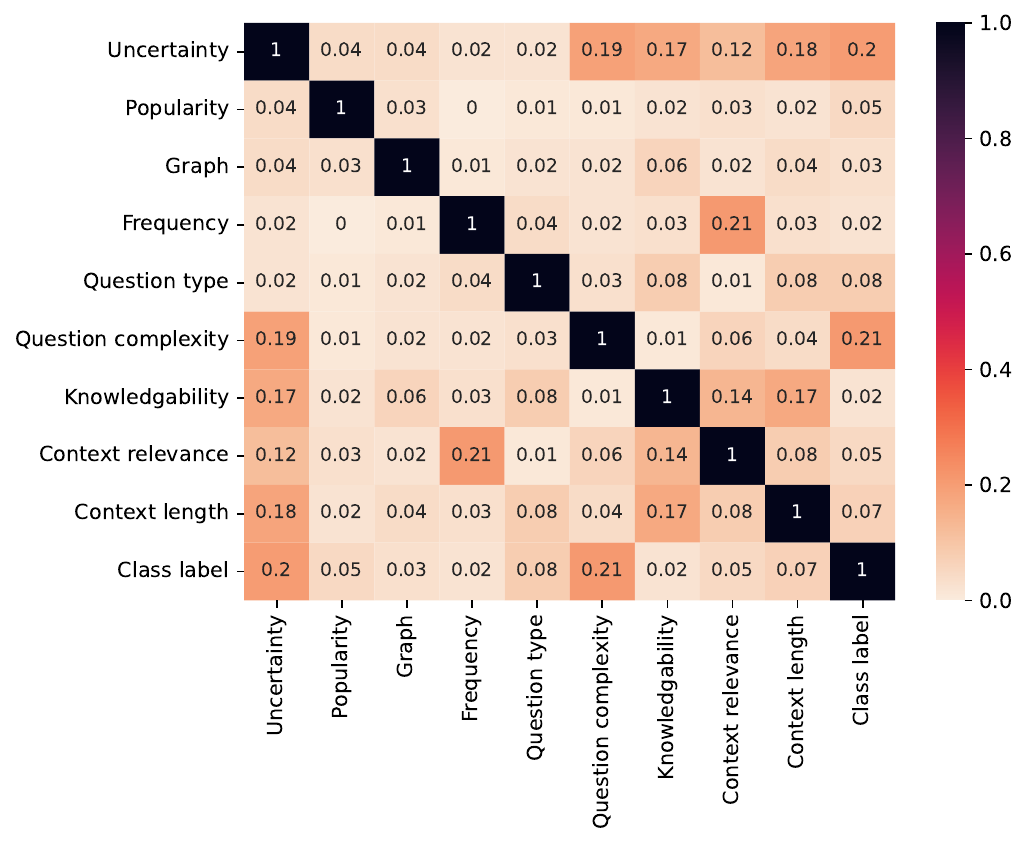}
  \caption{nq}
  \label{fig:1}
\end{subfigure}\hfil 
\begin{subfigure}{0.5\textwidth}
  \includegraphics[width=\linewidth]{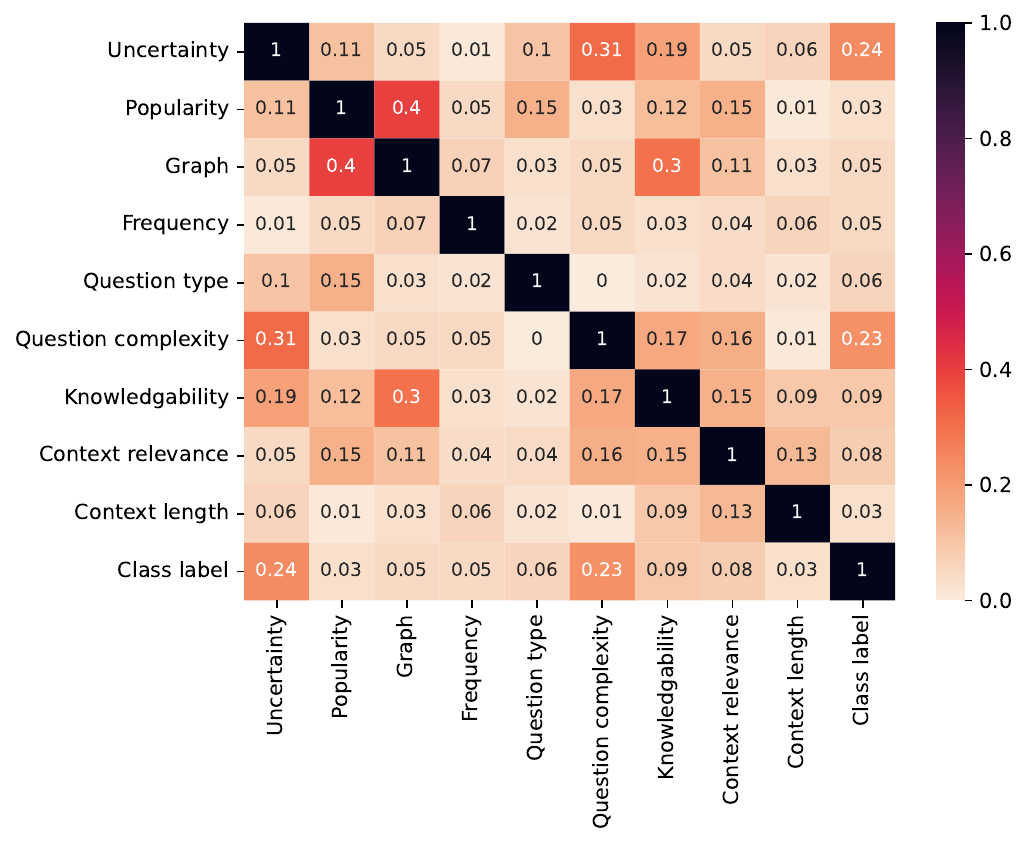}
  \caption{squad}
  \label{fig:2}
\end{subfigure}\hfil 

\medskip
\begin{subfigure}{0.5\textwidth}
  \includegraphics[width=\linewidth]{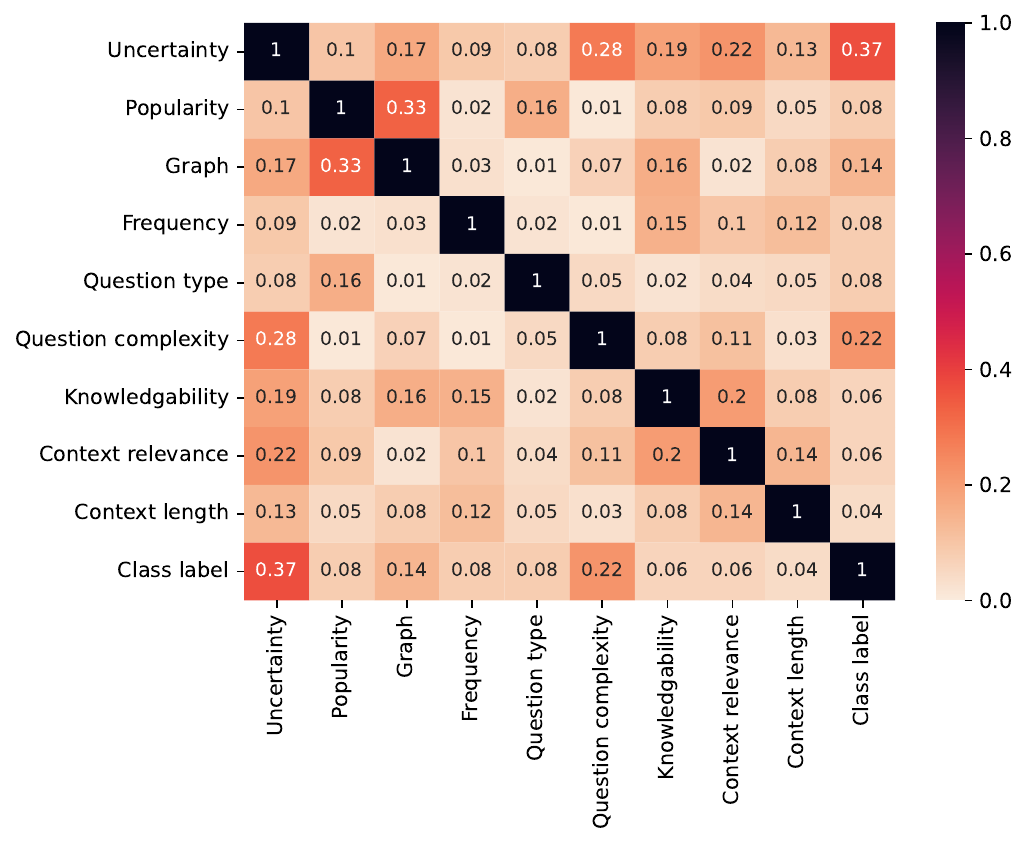}
  \caption{trivia}
  \label{Corr_trivia}
\end{subfigure}\hfil 
\begin{subfigure}{0.5\textwidth}
  \includegraphics[width=\linewidth]{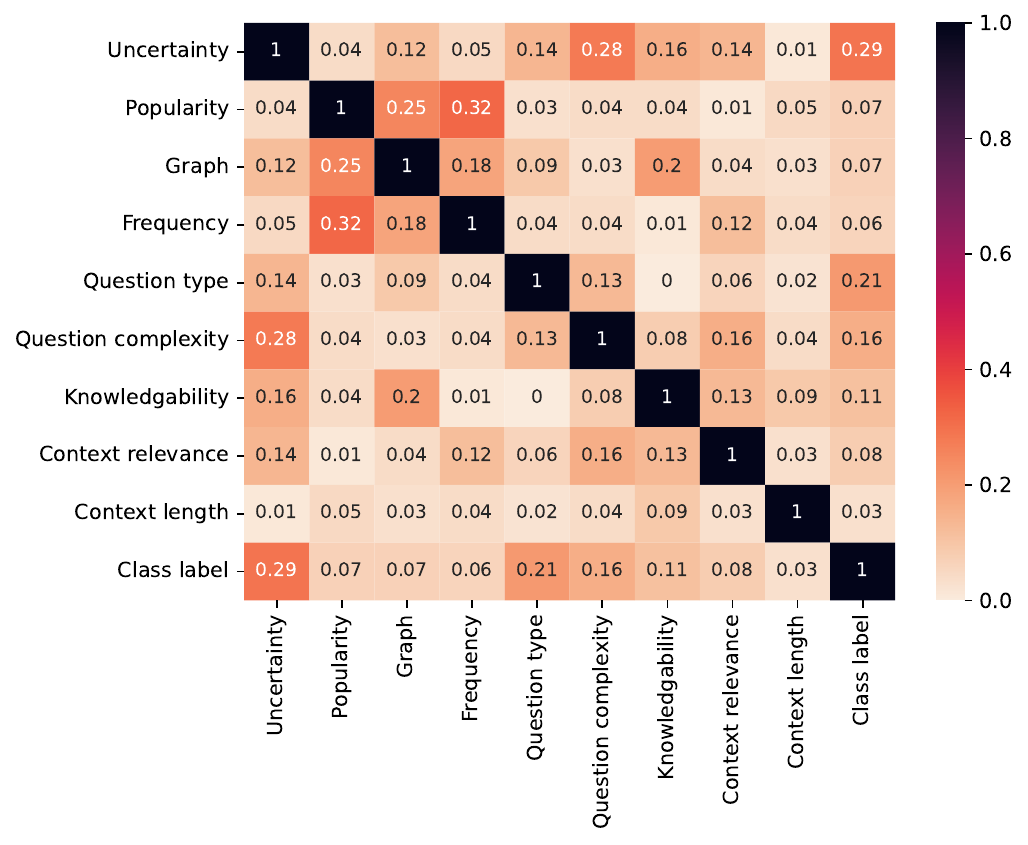}
  \caption{hotpot}
  \label{Corr_hotpot}
\end{subfigure}\hfil 

\medskip
\begin{subfigure}{0.5\textwidth}
  \includegraphics[width=\linewidth]{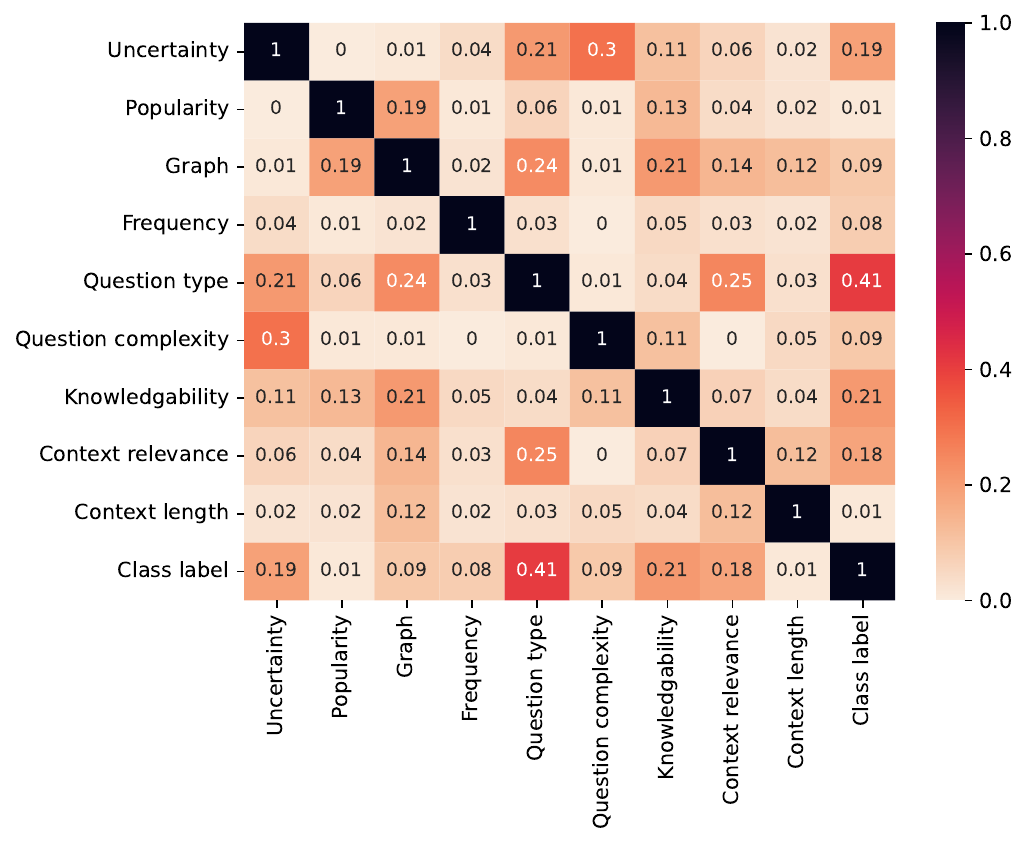}
  \caption{2wikimultihop}
  \label{Corr_2wiki}
\end{subfigure}\hfil 
\begin{subfigure}{0.5\textwidth}
  \includegraphics[width=\linewidth]{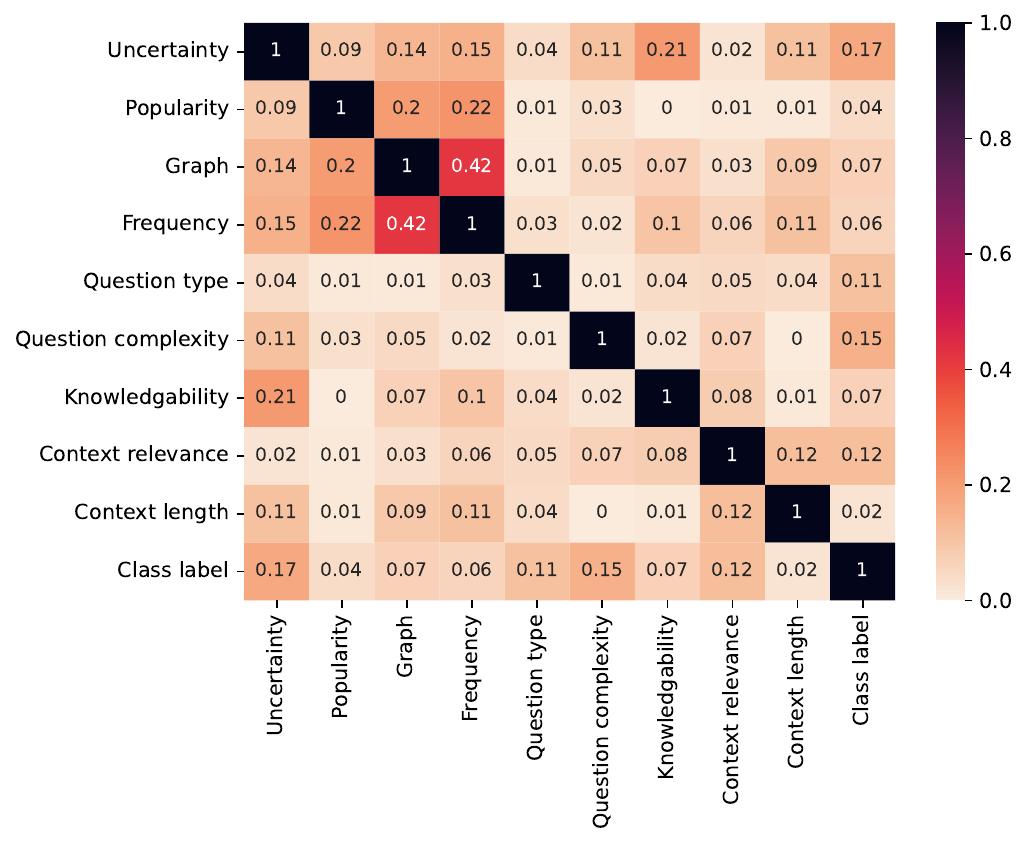}
  \caption{musique}
  \label{Corr_musique}
\end{subfigure}\hfil 

\caption{Absolute correlation of features from different groups of external features with class label}
\label{fig:corr_imgbyimg2}
\end{figure*}

\end{document}